\documentclass[runningheads]{llncs}
\usepackage{amsmath}
\usepackage{etoolbox}
\makeatletter
\patchcmd{\@biblabel}{#1.}{[#1]}{}{}
\makeatother
\usepackage{multirow}
\usepackage{bm}
\usepackage{subcaption}
\usepackage[T1]{fontenc}
\usepackage{hyperref}
\usepackage{amssymb}
\usepackage{rotating}
\usepackage{graphicx}

\usepackage[numbers,sort&compress]{natbib} 

\usepackage{etoolbox}
\makeatletter
\renewenvironment{thebibliography}[1]
  {\section*{\refname}%
   \small\list{\@biblabel{\@arabic\c@enumiv}}%
        {\settowidth\labelwidth{\@biblabel{#1}}%
         \leftmargin\labelwidth
         \advance\leftmargin\labelsep
         \@openbib@code
         \usecounter{enumiv}%
         \let\p@enumiv\@empty
         \renewcommand\theenumiv{\@arabic\c@enumiv}}%
   \sloppy\clubpenalty4000\widowpenalty4000%
   \sfcode`\.\@m}
  {\def\@noitemerr
    {\@latex@warning{Empty `thebibliography' environment}}%
   \endlist}
\renewcommand{\@biblabel}[1]{[#1]}
\makeatother

\begin{document}
\hypersetup{hypertex=true,
            colorlinks=true,
            linkcolor=blue,
            anchorcolor=blue,
            citecolor=blue}
\title{Concept Factorization via Self-Representation and Adaptive Graph Structure Learning}
\titlerunning{CFSRAG}
%




\author{
Zhengqin Yang\inst{1} \and
Di Wu\inst{2} \and
Jia Chen\inst{3} \and
Xin Luo\inst{2}
}

\authorrunning{Z. Yang et al.}

\institute{
Department of Computer Science and Technology, Chongqing University of Posts and Telecommunications, Chongqing, 400065, China \\
\email{yangzhengqin0@gmail.com}
\and
College of Computer and Information Science, Southwest University, Chongqing, 400715, China \\
\email{wudi.cigit@gmail.com, luoxin@swu.edu.cn}
\and
School of Cyber Science and Technology, Beihang University, Beijing, 100191, China \\
\email{chenjia@buaa.edu.cn}
}

\maketitle              
\begin{abstract}
Concept Factorization (CF) models have attracted widespread attention due to their excellent performance in data clustering. In recent years, many variant models based on CF have achieved great success in clustering by taking into account the internal geometric manifold structure of the dataset and using graph regularization techniques. However, their clustering performance depends greatly on the construction of the initial graph structure. In order to enable adaptive learning of the graph structure of the data, we propose a Concept Factorization Based on Self-Representation and Adaptive Graph Structure Learning (CFSRAG) Model. CFSRAG learns the affinity relationship between data through a self-representation method, and uses the learned affinity matrix to implement dynamic graph regularization constraints, thereby ensuring dynamic learning of the internal geometric structure of the data. Finally, we give the CFSRAG update rule and convergence analysis, and conduct comparative experiments on four real datasets. The results show that our model outperforms other state-of-the-art models. 

\end{abstract}
\section{Introduction}
Clustering is a fundamental task in unsupervised learning and remains one of the most actively explored research problems in data mining and machine learning. It aims to partition a dataset into groups (clusters) such that samples within the same group exhibit high similarity, thereby revealing intrinsic structures or hidden patterns in the data. Clustering techniques have found widespread applications across numerous domains, including image processing~\cite{b2}, text mining~\cite{b12}, biology~\cite{b4,b32}, and social network analysis~\cite{b3,b5,b9}. To improve clustering performance, a variety of approaches have been developed over the years, such as \textit{k}-means clustering~\cite{b65}, density peak clustering~\cite{b57,b87}, hierarchical clustering, spectral clustering, and non-negative matrix factorization (NMF)~\cite{b8}.Among them, NMF has gained significant attention due to its capability to yield part-based, interpretable representations by decomposing a non-negative data matrix $X \in \mathbb{R}^{M \times N}_+$ into two low-rank non-negative matrices $Q \in \mathbb{R}^{M \times f}_+$ and $V \in \mathbb{R}^{N \times f}_+$, such that $X \approx QV^T$. Here, $Q$ and $V$ are typically referred to as the basis matrix and the coefficient matrix, respectively. Owing to its non-negativity constraint, NMF avoids the cancellation of positive and negative components, making the resulting factors more interpretable in practical tasks. As a result, it has been widely applied in clustering analysis~\cite{b13,b14,b15,b16}, community detection~\cite{b6,b7,b79,b67}, feature extraction~\cite{b1,b22}, recommendation systems\cite{b48,b55,b56,b57,b58,b59,b10,b82,b83}, latent factor analysis~\cite{b49,b50,b51,b52,b53,b54,b81} and data recovery~\cite{b27,b33,b34,b35,b41,b42,b20,b39,b24}. Furthermore, NMF has shown great potential in large-scale data environments due to its ability to extract meaningful low-dimensional representations from high-dimensional inputs. However, as real-world data often contain missing entries or noise, especially in fields such as medical diagnosis, sensor networks, and social media, conventional NMF models face significant challenges when applied to incomplete datasets~\cite{b43,b44,b45,b46}. To address this issue, researchers have developed numerous variants of NMF tailored for ~\cite{b63,b64,b73,b74,b75,b76,b77}, enabling more robust performance in practical applications~\cite{b33,b35,b42}. In contrast to NMF, concept factorization (CF)~\cite{b12} was proposed to allow for the modeling of datasets containing negative entries. By expressing the basis matrix as a linear combination of data points in the original feature space, CF provides a sparse and flexible representation that extends the applicability of matrix factorization to a broader range of data types.
\par Although NMF and CF perform well in clustering, they still have some limitations. To address these, many models based on NMF and CF have been proposed over the years. For example, to preserve the local geometric structure between data samples, Cai et al.\cite{b13} proposed graph-regularized NMF (GNMF). Concurrently, based on CF and considering the nearest-neighbor graph of the data, they introduced locally consistent CF (LCCF)\cite{b14} for text clustering. Recognizing that outliers may cause large residuals when using Euclidean distance for error approximation, Chen and Li\cite{b15} proposed an entropy-minimization matrix factorization (EMMF) framework to address the impact of outliers. To make models more robust to noise and interference, Hu et al.\cite{b16} introduced a robust sparse CF framework (RSCF) for subspace learning. To learn the adaptive inherent graph structure of the data space while reducing the burden of explicit orthogonal constraints, Hu et al.\cite{b17} proposed Adaptive Graph Learning CF based on the Stiefel manifold (AGCF-SM). Moreover, to capture the complex global and local manifold structures within the data, Li et al.\cite{b18} introduced a dual graph regularized global and local CF model (DGLCF), which combines CF and graph regularization techniques.
\par The aforementioned graph-regularized models, such as GNMF, LCCF, and DGLCF, all share a common limitation: they construct a similarity graph from the dataset and then impose graph regularization constraints. This makes the process of similarity graph construction and the subsequent matrix factorization two independent steps. When dealing with complex datasets that contain significant noise, the constructed similarity graph may fail to accurately reflect the intrinsic manifold structure of the data, thereby severely impacting clustering performance. Hence, adaptive learning of the similarity graph becomes crucial. Although Hu et al. adopted a self-representation approach for adaptive similarity graph learning in AGCF-SM, the concept factorization and self-representation components were only superficially combined, lacking deeper integration. To address these shortcomings, this paper introduces a \textbf{C}oncept \textbf{F}actorization model with \textbf{S}elf-\textbf{R}epresentation and \textbf{A}daptive \textbf{G}raph Learning (\textbf{CFSRAG}). CFSRAG views concept factorization as a self-representation learning process, achieving adaptive learning of the similarity graph structure. Furthermore, it applies dynamic graph regularization via the learned similarity matrix, preserving the structural relationships between data samples. The main contributions of this paper are as follows:
\begin{itemize}
	\item We propose a novel model, CFSRAG, which adaptively learns the similarity relationships between data samples while preserving their local geometric structures using both the similarity graph matrix and feature manifold information.
    \item We provide the multiplicative update rules for our model.
    \item Extensive comparative experiments demonstrate that our model outperforms state-of-the-art methods.
\end{itemize}

\section{Preliminaries}
\section{Preliminary}
\subsection{NMF}
Given a data matrix $X = [x_1, x_2, \dots, x_n] \in \mathbb{R}^{m \times n}$ , where m represents the dimensionality of the features, and n denotes the number of samples, the NMF model decomposes $X$ into two low-rank matrices $Q \in R _ { + } ^ { m \times c }$ and $  V \in R _ { + } ^ { n \times c }$ , where $c$ corresponds to the number of clusters, $Q$ represents the basis matrix composed of $c$ cluster centroids, and $V$ serves as indicator matrix of the samples. The objective function of NMF is formulated as
\begin{equation}
\min_{U \geq 0, V \geq 0} \left\| X - QV^T \right\|_F^2.
\end{equation}
For problem (1), the corresponding update rule is:
\begin{equation}
q_{ij} \leftarrow q_{ij} \frac{(X V)_{i j}}{\left(Q V^{T} V\right)_{i j}}, v_{i j} \leftarrow v_{i j} \frac{\left(X^{T} Q\right)_{i j}}{\left(V Q^{T} Q\right)_{i j}}.
\end{equation}
\subsection{Concept Factorization}
To enhance the interpretability of the model and address the non-negativity of data, Xu and Gong[12] proposed the CF model. In the CF model, the non-negative linear combination $XU$ of the data matrix $X$ is used to replace the basis matrix $Q$ in the NMF model. The objective function of CF is formulated as
\begin{equation}
 \min _ { U \geq 0 , V \geq 0 } | | X - X U V ^ { T } | | _ { F } ^ { 2 } .
\end{equation}
Xu and Gong\cite{b12} provided the following multiplicative update rules for Equation (3):
\begin{equation}
u_{ij} \leftarrow u_{ij} \frac{(KV)_{ij}}{\left(KUV^{T} V\right)_{ij}}, v_{ij} \leftarrow v_{i j} \frac{\left(KU\right)_{i j}}{\left(VU^{T}KU\right)_{i j}}
\end{equation}
where $K=X^{T}X$.
\section{Methods}
\subsection{Construct the Initial Affinity Matrix}
The initial affinity matrix is constructed based on the original feature data $X$. When two data points are more similar, there should be a stronger affinity between them. To obtain the optimal affinity matrix, Nie et al.\cite{b19} proposed the following objective function:
\begin{equation}
\begin{aligned}
\underset{A}{\min}& \sum_{i,j}^n \left( \lVert x_i - x_j \rVert^2 a_{ij} + \lambda \lVert A \rVert_F^2 \right), \\
\text{s.t.}  & \lVert a_i \rVert =1,0\le a_i\le 1, 
\end{aligned}
\end{equation}
where $a_i$ denotes the i-th column of $A$, and $\lambda$ is a balancing parameter used to weigh the trade-off between the first and second terms of the objective function. The inclusion of the second term helps prevent the trivial solution $A=I$, ensuring that the learned affinity matrix is meaningful and not merely the identity matrix.
\par Define the distance matrix $D'$  such that $d_{ij}^{'}=\lVert x_i-x_j \rVert ^2$, and $d_{i}^{'}$  denotes the i-th column of $D'$. The (5) can be decoupled into subproblems for each column of $A$, which can be formulated as follows:
\begin{equation}
 \underset{\lVert a_i \rVert =1,0\le a_i\le 1}{\min}\lVert a_i+\frac{1}{2\gamma}d_{i}^{'} \rVert _{2}^{2}.  
\end{equation}
For problem (6), Nie et al.\cite{b19} provided the following closed-form solution:
\begin{equation}
a_{ij}=\frac{\left( d_{i\left( p+1 \right)}^{''}-d_{ij}^{''} \right) _+}{pd_{i\left( p+1 \right)}^{''}-\sum_u^p{d_{iu}^{''}}},
\end{equation}
where  $d_{i}^{''}$ is the result of sorting $d_{i}^{'}$  in ascending order, $p$ denotes the number of nearest neighbors to retain with the largest weights, and $\left( \cdot \right) _+$
indicates $\max \left( \cdot ,0 \right) $. The weight matrix $W$ is given by $W=(A+A^T)/2$.
\subsection{CFSRAG}
The traditional CF model not only inherits the interpretability advantages of NMF but also imposes no constraints on the input data. In the CF model, $X=XUV^T$ , and by incorporating the concept of data self-representation and treating $UV^T$  as a unified entity, the product $UV^T$ is replaced by an affinity matrix $Z$. As a result, the CF model transforms into $X=XZ$ , becoming a self-representation model. According to the Euclidean distance adopted in \cite{b21,b23,b25,b26,b29,b30,b37} , the specific objective function is as follows:
\begin{equation}
\begin{aligned}
F_{ CFSRAG }&=||X-XZ||_ {F}^{2}+\alpha|| Z - U V ^ { T }||_{F}^ {2 } + \beta T r ( V ^ { T } L V ) 
\\&+ \lambda || Z | | _ { F } ^ { 2 } , \\
\text{s.t.}  &Z \geq 0 , U \geq 0 , V \geq 0 
\end{aligned}
\end{equation}
where, $\alpha >0$ and $\beta >0$  represent the regularization parameters used to balance the dynamic reconstruction affinity matrix term and the graph regularization term, respectively, and  $\lambda >0$ represents the parameter used to control the weight of  $\||Z||_{F}^{2}$ to prevent the learning of a trivial solution of $Z=I$.  $L$ denotes the Laplacian matrix of the affinity matrix $Z$, defined as $L=D-W$, where $W$ is the initial weight matrix generated in Section $A$ and $D$ is a diagonal matrix of the weight matrix $W$ whose element $d_{ii}=\sum_j{w_{ij}}$
 . For ease of subsequent derivations, let $\lVert A \rVert _{F}^{2}=Tr\left( A^TA \right)$, where $\text{Tr}\left( \cdot \right) $
 denotes the trace of a matrix. We transform (8) into:
\begin{equation}
\begin{aligned}
F_{CFSRAG} &= \ \text{Tr}(K - 2KZ + Z^T KZ) + \beta \text{Tr}(V^T LV) \\
&+ \alpha \text{Tr}(Z^T Z - 2Z^T UV^T + VU^T UV^T) + \lambda r(Z^T Z) \\
\text{s.t.} &\ Z \geq 0, U \geq 0, V \geq 0
\end{aligned}
\end{equation}
Let \(\Gamma = [\gamma_{ik}]\), \(\Psi = [\varphi_{jk}]\), and \(\Phi = [\phi_{ij}]\) represent the Lagrange multipliers for the non-negativity constraint \(U = [u_{ik}] > 0\), \(V = [v_{jk}] > 0\), and \(Z = [z_{ij}] > 0\), respectively. The Lagrangian function is then defined as follows:
\begin{equation}
\begin{aligned}
L_{CFSRAG} &= \ \text{Tr}(K - 2KZ + Z^T KZ) + \beta \text{Tr}(V^T LV) \\
&+ \alpha \text{Tr}(Z^T Z - 2Z^T UV^T + VU^T UV^T) + \lambda r(Z^T Z) \\&+\text{Tr}(\Gamma U^T)+\text{Tr}(\Psi V^T)+\text{Tr}(\Phi Z^T).
\end{aligned}
\end{equation}
Taking the partial derivatives of the Lagrangian function $L_{CFSRAG}$  with respect to $U$ and $V$ and setting it to zero, we have
\begin{equation}
\begin{aligned}
&-2ZV+2UV^{T}V+\Gamma=0, \\&-2\alpha Z^TU + 2 \alpha V U ^ { T } U + 2 \beta D V - 2 \beta W V + \Psi = 0 
\end{aligned}
\end{equation}
When taking the partial derivative of $Z$, the trace term  $\text{Tr}(V^TLV)$ can be written as $\sum_{i,j}^n \lVert v_i - v_j \rVert^2 z_{ij}$. By defining $H_{ij}=||v_{i}-v_{j}||^{2}$ , where  represents the i-th row of matrix $V$, the trace expression simplifies to $\text{Tr}(H Z^T)$. At this point, we have
\begin{equation}
\begin{aligned}
-2K + 2KZ + 2\alpha Z - 2\alpha UV^{T} + \beta H + 2\lambda Z + \Phi = 0.
\end{aligned}
\end{equation}
Then based on the KKT conditions  $\gamma_{ik} u_{ik} = 0$, $\varphi_{jk} v_{jk} = 0$ and $\phi_{ij} z_{ij} = 0$, we obtain
\begin{equation}
\begin{aligned}
&-(ZV)_{ik} u_{ik} + (UV^{T} V)_{ik} u_{ik}=0, \\
&-(\alpha Z^{T} U + \beta WV)_{jk} v_{jk} + (\alpha V U^{T} U + \beta D V)_{jk} v_{jk}=0, \\
&-(K + \alpha UV^{T})_{ij} z_{ij} + (KZ + (\alpha + \lambda)Z + \frac{1}{2} \beta H)_{ij} z_{ij}=0.
\end{aligned}
\end{equation}
Thus, we can derive the multiplicative update rule for $u_{ik}$, $v_{jk}$ and $z_{ij}$ as follows:
\begin{equation}
\begin{aligned}
u_{ik}\gets u_{ik}\frac{\left( ZV \right) _{ik}}{\left( UV^TV \right) _{ik}},
\end{aligned}
\end{equation}
\begin{equation}
\begin{aligned}
v_{jk}\gets v_{jk}\frac{\left( \alpha Z^TU+\beta WV \right) _{jk}}{\left( \alpha VU^TU+\beta DV \right) _{jk}},
\end{aligned}
\end{equation}
\begin{equation}
\begin{aligned}
z_{ij}\gets z_{ij}\frac{\left( K+\alpha UV^T \right) _{ij}}{\left( KZ+\left( \alpha +\lambda \right) Z+\dfrac{1}{2}\beta H \right) _{ij}}.
\end{aligned}
\end{equation}

\section{Experiments}
\subsection{General Settings Evaluation Metric}
In the experiments of this paper, we adopt the widely used normalized mutual information (NMI), accuracy (ACC), and purity (PUR) to evaluate the model's performance.
\begin{table}[ht]
\centering
\caption{Statistical Description of Datasets.}
\setlength{\tabcolsep}{8pt} 
    \renewcommand{\arraystretch}{1.4} 
\begin{tabular}{c c c c c}
\hline

\hline
ID & Dataset & Size (n) & Dimensionality (m) & Classes (c) \\
\hline
D1 & zoo & 101 & 16 & 7 \\
D2 & JAFFE & 213 & 4096 & 10 \\
D3 & ORL & 400 & 1024 & 40 \\
D4 & YALE & 165 & 1024 & 15 \\
\hline

\hline
\end{tabular}
\label{table1}
\end{table}

\begin{sidewaystable*}[htbp]
    \centering
    \caption{Clustering Performance ((NMI, ACC, PUR)\%$\pm$STD\%) of Different Models on Different Data, Including Win/Loss Counts, Signed‐Ranks Test and Friedman Test.}
    \setlength{\tabcolsep}{3pt} 
    \renewcommand{\arraystretch}{1.4} 

    \begin{tabular}{c c c c c c c c c c c} 
        \hline

\hline
        \textbf{Metric} & \textbf{Dataset} & \textbf{M1} & \textbf{M2}& \textbf{M3}& \textbf{M4}& \textbf{M5} & \textbf{M6}& \textbf{M7}& \textbf{M8}& \textbf{CFSRAG}\\ 
        \hline

\hline
        \multirow{4}{*}{\textbf{NMI}} & \textbf{D1} & 65.85$\pm$8.76 & 62.89$\pm$5.26 & 69.07$\pm$5.07 & 69.45$\pm$4.03 & 75.34$\pm$4.4 & 73.16$\pm$2.65 & 80.61$\pm$2.23 & 76.67$\pm$2.12 & \textbf{87.08$\pm$0.97} \\
        & \textbf{D2} & 71.97$\pm$5.73 & 71.56$\pm$5.62 & 78.73$\pm$3.15 & 81.57$\pm$3.52 & 74.39$\pm$3.07 & 82.51$\pm$3 & 82.31$\pm$3.55 & 88.88$\pm$3.18 & \textbf{95.52$\pm$2.05} \\
        & \textbf{D3} & 66.4$\pm$1.25 & 67.19$\pm$0.93 & 67.37$\pm$0.77 & 79.69$\pm$1.61 & 74.11$\pm$1.94 & 80.16$\pm$1.54 & 82.76$\pm$1.03 & 81.93$\pm$0.77 & \textbf{89.41$\pm$1.28} \\
        & \textbf{D4} & 59.53$\pm$1.82 & 62.35$\pm$2.36 & 60.56$\pm$1.66 & 66.22$\pm$2.13 & 66.03$\pm$2.14 & 67.18$\pm$1.91 & 66.99$\pm$1.2 & 67.46$\pm$1.44 & \textbf{74$\pm$2.31} \\
        \hline
        \multirow{4}{*}{\textbf{ACC}} & \textbf{D1} & 73.37$\pm$10.17 & 64.55$\pm$5.01 & 71.78$\pm$6.27 & 68.42$\pm$6.97 & 68.71$\pm$10.28 & 71.49$\pm$5.7 & 81.88$\pm$3.96 & 79.7$\pm$1.78 & \textbf{87.03$\pm$0.82} \\
        & \textbf{D2} & 68.36$\pm$5.7 & 66.95$\pm$7.98 & 68.54$\pm$5.87 & 76.67$\pm$5.74 & 69.62$\pm$5.12 & 77.98$\pm$5.4 & 78.54$\pm$6.21 & 82.91$\pm$7.44 & \textbf{88.22$\pm$5.78} \\
        & \textbf{D3} & 44.2$\pm$2 & 44.47$\pm$1.97 & 44.72$\pm$1.77 & 60.68$\pm$3.17 & 54.72$\pm$4.76 & 61.42$\pm$3.12 & 64.88$\pm$2.76 & 63.73$\pm$2.21 & \textbf{78.78$\pm$4.26} \\
        & \textbf{D4} & 48.61$\pm$3.34 & 53.21$\pm$3.15 & 50.36$\pm$4.01 & 60.85$\pm$4.03 & 60.55$\pm$2.96 & 62.18$\pm$4.11 & 60.42$\pm$2.89 & 61.21$\pm$3.34 & \textbf{64.55$\pm$3.55} \\
        \hline
        \multirow{4}{*}{\textbf{PUR}} & \textbf{D1} & 78.81$\pm$5.11 & 77.23$\pm$4.01 & 78.71$\pm$3.01 & 83.96$\pm$2.54 & 85.64$\pm$2.31 & 84.95$\pm$2.16 & 87.03$\pm$3.02 & 79.9$\pm$1.72 & \textbf{90.4$\pm$0.89} \\
        & \textbf{D2} & 70.42$\pm$5.06 & 68.97$\pm$7.39 & 70.7$\pm$5.77 & 79.39$\pm$4.13 & 71.97$\pm$4.73 & 80.89$\pm$3.48 & 80.38$\pm$4.84 & 84.79$\pm$5.75 & \textbf{88.92$\pm$5.46} \\
        & \textbf{D3} & 46.62$\pm$1.79 & 46.92$\pm$1.77 & 47.37$\pm$1.67 & 65.22$\pm$2.47 & 57.55$\pm$4.44 & 66$\pm$2.4 & 69.4$\pm$2.21 & 67.33$\pm$1.91 & \textbf{81.15$\pm$4.16} \\
        & \textbf{D4} & 50.61$\pm$3.15 & 55.76$\pm$3.33 & 52.24$\pm$3.94 & 61.88$\pm$3.98 & 61.7$\pm$2.98 & 63.03$\pm$4 & 61.58$\pm$2.6 & 62.06$\pm$3.45 & \textbf{66.24$\pm$3.6} \\
        \hline

\hline
        \multirow{3}{*}{\textbf{Statistic}} & \textbf{win/loss} & 12/0 & 12/0 & 12/0 & 12/0 & 12/0 & 12/0 & 12/0 & 12/0 & \textbf{96/0*} \\
        & \textbf{F-rank} & 8.08 & 8.25 & 7.08 & 5.17 & 5.58 & 3.67 & 3.25 & 2.92 & \textbf{1.0} \\
        & \textbf{p-value} & 5.32$\times$10$^{-4}$ & 5.32$\times$10$^{-4}$ & 5.32$\times$10$^{-4}$ & 5.32$\times$10$^{-4}$ & 5.32$\times$10$^{-4}$ & 5.32$\times$10$^{-4}$ & 5.32$\times$10$^{-4}$ & 5.32$\times$10$^{-4}$ & - \\
       \hline

\hline  
        \multicolumn{10}{l}{* The total win/loss cases of CFSRAG} \\
    \end{tabular}
    \label{table4}
\end{sidewaystable*}

\textbf{Datasets}: We used four widely employed real-world datasets in clustering research, including image datas and UCI dataset. The specific statistical details of these datasets are presented in Table~\ref{table1}. 

\textbf{Tested Models}: We evaluated the performance of our model CFSRAG together with NMF, EMMF, GNMF, CF, RSCF, LCCF, AGCF\_SM and DGLCF. We renumbered the comparison models from M1 to M8.

\subsection{Performance Comparison of Different Models}
In order to illustrate the effectiveness of the CFSRAG model in clustering, we conducted a comparative experiment with our model and the comparison model, and calculated the mean and standard deviation results of NMI, ACC and PUR respectively. The detailed statistical results are listed in Table~\ref{table4}. Experimental results show that CFSRAG achieves optimal performance on all datasets, significantly outperforming baseline models and other advanced methods. This advantage is attributed to the self-representation learning and dynamic graph regularization methods integrated in the CFSRAG model design, which enable the model to more effectively capture the intrinsic structure of the data. In particular, compared with the M7, which also adopts self-representation learning, CFSRAG ensures more accurate learning of graph structure information by further decomposing the affinity matrix $Z$ and establishing a connection with the indicator matrix $V$, resulting in better performance on all data sets Both are better than M7. In order to verify the statistical significance of the performance improvement of the CFSRAG model, we performed Friedman test\cite{b28,b60,b61,b69}, win/loss\cite{b62,b68,b70} count and signed rank test\cite{b31}. The results show that our model achieved the lowest value (1.0) in the F-rank value, confirming the significant advantage of CFSRAG in clustering performance, and the results of the signed rank test show that the performance difference between our model and the comparison model is statistically significant ($p$-value $<$ 0.05). Finally, in the win/loss count, the CFSRAG model achieved an overwhelming victory of 96/0 on all datasets, further emphasizing its superior performance in clustering tasks. This study experimentally verified the effectiveness of the CFSRAG model in clustering tasks, and significantly improved the clustering performance by combining self-representation learning with dynamic graph regularization, providing strong evidence support for the potential of the CFSRAG model in practical applications.

\subsection{Ablation analysis}
To demonstrate the effectiveness of the self-representation learning graph structure and dynamic graph regularization employed in the CFSRAG model, we conducted the following ablation experiments. We compared CFSRAG with the following variant models:

a) CFSR: A variant model based on CFSRAG, with $\beta=0$ and $\gamma=0$, which utilizes only self-representation learning.

b) CFSR-F: A variant model based on CFSRAG, with $\beta=0$, which adds an F-norm constraint on the affinity matrix $Z$ to the CFSR model.

c) CFSRG: A variant model based on CFSRAG that uses the initially computed Laplacian matrix and does not update the Laplacian matrix during the model iteration process
The experimental comparison results are shown in Table~\ref{table5}.

\begin{table}[h]
\centering
\caption{Comparison of the Performance between CFSRAG and the Ablation Contrast Models CFSR, CFSR-F and CFSRG,Including Win/Loss Counts, Signed-Ranks Test and Friedman Test.}
    \setlength{\tabcolsep}{2pt} 
        \renewcommand{\arraystretch}{1.4} 

\begin{tabular}{cccccc}
\hline

\hline
\textbf{Metrics} & \textbf{Dataset} & \textbf{CFSR} & \textbf{CFSR-F} & \textbf{CFSRG} & \textbf{CFSRAG} \\
\hline

\hline
\multirow{4}{*}{\textbf{NMI}} & D1 & $77.78_{\pm 2.62}$ & $83.26_{\pm 3.32}$ & $86.82_{\pm 1.44}$ & $\mathbf{87.08_{\pm 0.97}}$ \\
 & D2 & $84.77_{\pm 2.97}$ & $84.81_{\pm 2.97}$ & $94.04_{\pm 1.96}$ & $\mathbf{95.52_{\pm 2.05}}$ \\
& D3 & $79.85_{\pm 1.11}$ & $86.09_{\pm 0.99}$ & $88.69_{\pm 1.04}$ & $\mathbf{89.41_{\pm 1.28}}$ \\
& D4 & $64.97_{\pm 2.94}$ & $67.38_{\pm 2.12}$ & $72.02_{\pm 1.8}$ & $\mathbf{74_{\pm 2.31}}$ \\
\hline
\multirow{4}{*}{\textbf{ACC}} & D1 & $77.92_{\pm 4.91}$ & $81.49_{\pm 6.26}$ & $86.73_{\pm 0.91}$ & $\mathbf{87.03_{\pm 0.82}}$ \\
 & D2 & $73.99_{\pm 4.81}$ & $73.99_{\pm 4.81}$ & $\mathbf{91.55_{\pm 4.91}}$ & $88.22_{\pm 5.78}$ \\
& D3 & $53.23_{\pm 1.76}$ & $71.15_{\pm 2.07}$ & $77.9_{\pm 2.42}$ & $\mathbf{78.78_{\pm 4.26}}$ \\
& D4 & $51.82_{\pm 3.68}$ & $53.88_{\pm 4}$ & $\mathbf{65.58_{\pm 3.2}}$ & $64.55_{\pm 3.55}$ \\
\hline
\multirow{4}{*}{\textbf{PUR}} & D1 & $86.53_{\pm 2.22}$ & $89.31_{\pm 2.38}$ & $\mathbf{90.69_{\pm 0.91}}$ & $90.4_{\pm 0.89}$ \\
 & D2 & $76.34_{\pm 4.32}$ & $76.34_{\pm 4.32}$ & $\mathbf{92.25_{\pm 4.13}}$ & $88.92_{\pm 5.46}$ \\
& D3 & $59.2_{\pm 1.48}$ & $75.25_{\pm 1.74}$ & $80.73_{\pm 1.9}$ & $\mathbf{81.15_{\pm 4.16}}$ \\
& D4 & $54.97_{\pm 3.34}$ & $56.55_{\pm 2.78}$ & $\mathbf{66.36_{\pm 3.03}}$ & $66.24_{\pm 3.6}$ \\
\hline
\multirow{3}{*}{\textbf{Statistic}} & \textbf{win/loss} & 12/0 & 12/0 & 7/5 & \textbf{31/5*} \\
        & \textbf{F-rank} & 3.92 & 3.08 & 1.58 & \textbf{1.42} \\
        & \textbf{p-value} & 5.32$\times$10$^{-5}$ & 5.32$\times$10$^{-5}$ & 5.64$\times$10$^{-1}$ & - \\
\hline

\hline
\end{tabular}
\label{table5}
\end{table}

\begin{table}[ht]
    \centering
    \caption{Wilcoxon Signed Rank Test Results for Different Ablation Comparison Models(Significance Level: $\alpha=0.05$)}
    \setlength{\tabcolsep}{12pt} 
        \renewcommand{\arraystretch}{1.4} 
    \begin{tabular}{cccc}
    \hline

    \hline
    \textbf{Comparison} & \textbf{R+}  &  \textbf{R-} & \textbf{p-value}\\
    \hline

    \hline
    CFSR-F vs CFSR  & 76.5 &  1.5 & 3.26$\times$10$^{-3}$ \\
    CFSRG vs CFSR & 78.0 &  0.0 & 2.22$\times$10$^{-3}$ \\
    CFSRAG vs CFSR & 78.0 &  0.0 & 2.22$\times$10$^{-3}$ \\
    CFSRG vs CFSR-F & 78.0 &  0.0 & 2.22$\times$10$^{-3}$ \\
    CFSRAG vs CFSR-F & 78.0 &  0.0 & 2.22$\times$10$^{-3}$ \\
    CFSRAG vs CFSRG & 43.0 &  35.0 & 7.54$\times$10$^{-1}$ \\
    \hline

    \hline
    \end{tabular}
    \label{table6}
\end{table}

The experimental comparison results are shown in Table~\ref{table5}, and the Wilcoxon signed-rank test results for the data in Table~\ref{table5} are summarized in Table~\ref{table6}. From these tables, we can draw the following conclusions:

a) The CFSR model based only on self-representation learning performs significantly lower than other variant models in terms of NMI, ACC, and PUR, and performs the worst in all datasets. This shows that although self-representation learning can capture part of the structural information of the data, its overall performance is limited due to the lack of regularization of the self-representation matrix and the limited ability to dynamically model local structure.

b) The CFSR-F model incorporating F-norm constraints shows significant improvements in all metrics compared to CFSR. For example, on the D3 dataset, NMI increases from 79.85 to 86.09, ACC increases from 53.23 to 71.15, and PUR increases from 59.20 to 75.25. By performing the Wilcoxon signed-rank test \cite{b72, b85} on CFSR-F and CFSR, we found that the performance difference between the models was statistically significant, $R+$ = 76.5, $R-$ = 1.5, $p$-value = 3.26$\times10^{-3}$. These results verify that the F-norm constraint enhances the sparsity and discriminative ability of the self-representation matrix $Z$, allowing the model to better adapt to the complex distribution of the data.

c) The CFSRG model that combines static graph regularization with F-norm regularized self-representation shows further improvements on most datasets, especially on the more complex datasets D2 and D3. For example, on the D2 dataset, NMI increases from 84.81 (CFSR-F) to 94.04, ACC from 73.99 to 91.55, and PUR from 76.34 to 92.25. The Wilcoxon signed rank test was performed on CFSRG and CFSR-F, $R+$ = 78.0, $R-$ = 0.0, $p$-value = 2.22$\times10^{-3}$. This shows that by capturing local relationships between data points, graph regularization significantly enhances the model's ability to model the global and local structure of the data.

\begin{figure}[!t]
    \centering
    \includegraphics[width=0.8\linewidth]{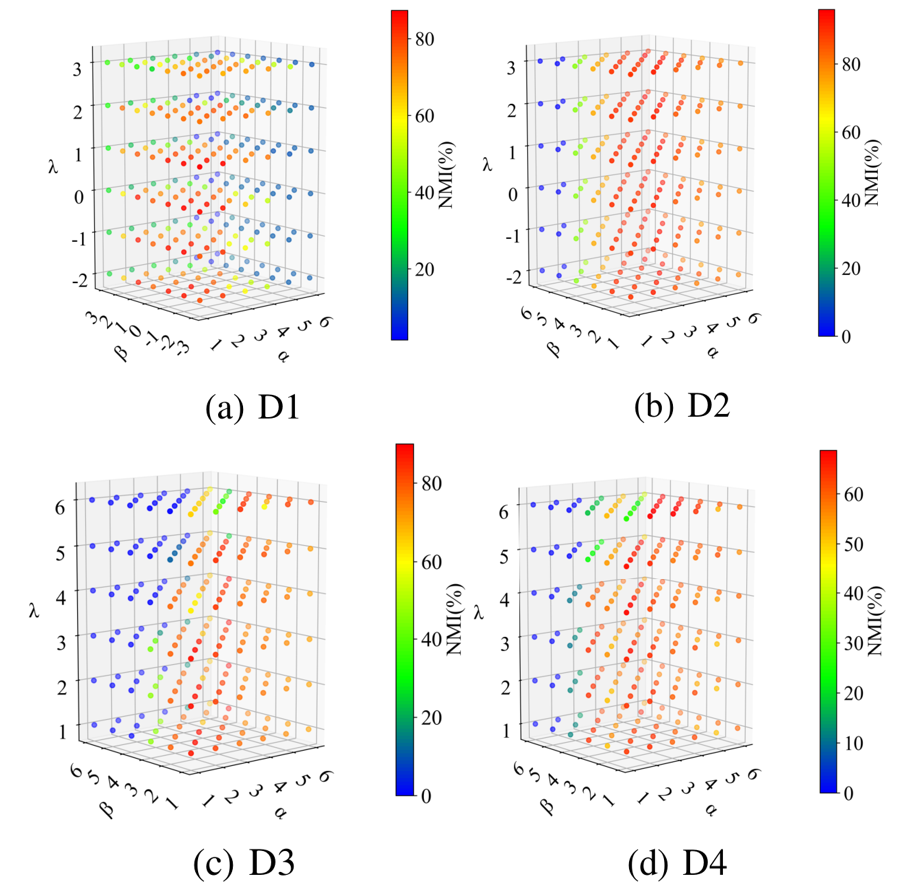} 
    \caption{CFSRAG clustering performance changes on D1, D2, D3 and D4, respectively, when $\alpha$, $\beta$ and $\lambda$ change.}
    \label{fig1}
\end{figure}


\begin{figure}[!t]
    \centering
    \includegraphics[width=0.8\linewidth]{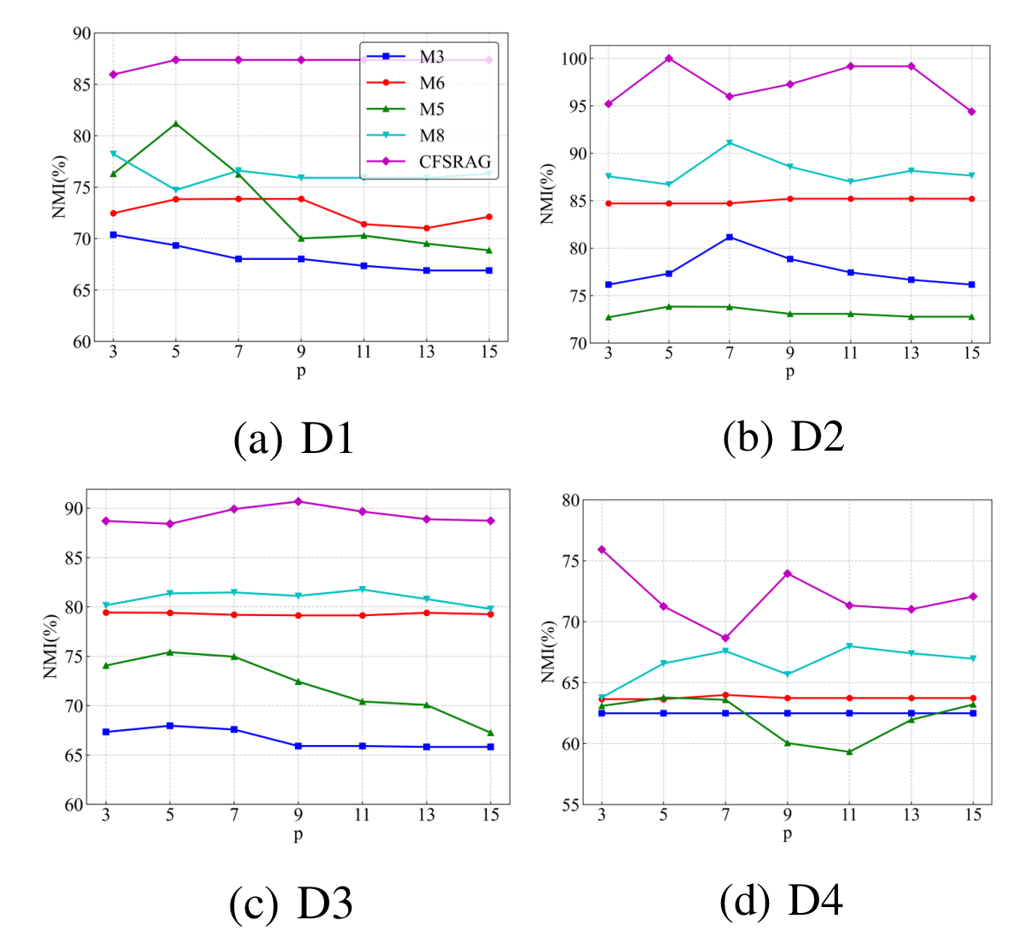}
    \caption{The variation in clustering performance of different models on D1, D2, D3 and D4 as the number of nearest neighbors $p$ changes.}
    \label{fig2}
\end{figure}

d) The full model CFSRAG, which combines dynamic graph regularization with F-norm regularized self-representation, achieves the best or near-best performance on all datasets. For example, on the D3 dataset, NMI increases from 88.69 (CFSRG) to 89.41, ACC increases from 77.9 to 78.78, and PUR increases from 80.73 to 81.15. This shows that by leveraging dynamic structural relationships learned through self-representation, the model can impose dynamic constraints on local geometric relationships between data points, further improving performance. Perform Wilcoxon signed rank test on CFSRAG and CFSRG, $R+$ = 43.0, $R-$ = 35.0, $p$-value = 0.754. Although this relatively high $p$-value comparison lacks statistical significance, CFSRAG outperforms CFSRG on 7 of the 12 indicators, showing a good trend. This suggests that dynamic graph regularization may provide real improvements that are not fully captured by current statistical analyses, perhaps due to the limited number of data sets or variability within the data.

\subsection{Hyper-parameter Analysis}
The hyperparameter tuning of the CFSRAG model has a significant impact on its overall performance. In this section, we conducted sensitivity experiments on three key hyperparameters of the CFSRAG model: $\alpha$, $\beta$ and $\lambda$. We evaluated the model's performance under various combinations of these hyperparameters and plotted 3D graphs, as shown in Fig.~\ref{fig1}. For brevity, we report only the influence on NMI values. It can be seen from the Fig.~\ref{fig1} that our model is very sensitive to the values of the hyperparameters    $\alpha$, $\beta$ and $\lambda$, and the performance of the model is affected by different parameter combinations. Therefore, it is particularly important to choose different parameter combinations for different data.

\par The parameter $p$ represents the choice of the number of neighbors when constructing the nearest neighbor graph, which significantly impacts the model's clustering performance. In the comparative models, M3, M5, M6 and M8 include the parameter $p$. We compare the clustering performance of CFSRAG with these models under different neighbor counts. The range of values for ppp across the four datasets is set uniformly to [3, 5, 7, 9, 11, 13, 15]. The experimental results are illustrated in Fig.~\ref{fig2}. From the figure, we can see that our model consistently achieves optimal performance across all values of $p$. Additionally, it is noted that the choice of  $p$ impacts different models to varying degrees across different datasets. Our CFSRAG model stabilizes when $p$ is greater than or equal to 5 on the D1 dataset. In contrast, optimal performance is achieved at $p$ values of 5, 9 and 3 for the D2, D3 and D4 datasets, respectively.

\section{Conclusion}
In this paper, we propose the CFSRAG model, building upon traditional concept factorization by integrating self-representation learning and graph regularization constraints. The model is based on two main ideas: (a) Dynamically learning the affinity relationships between samples using a self-representation method; and (b) Employing the learned affinity matrix to apply graph regularization to the clustering indicator matrix, thereby preserving the local geometric structure of the samples. Finally, experiments on four real-world datasets demonstrate that, compared to state-of-the-art models, CFSRAG exhibits superior performance in data clustering. It is worth noting that the model proposed in this paper contains many weight parameters, and particle swarm optimization\cite{b11,b38,b78} algorithm can be considered for adaptation in future research, and the algorithm includes a large number of matrix calculations, so parallel computing\cite{b36,b47} can be used to improve the calculation efficiency.  In addition, other distance measurement methods in \cite{b66} can be used to replace European distance.

\end{document}